\documentclass[sigconf]{acmart}

\usepackage{tablefootnote}
\usepackage{algpseudocode}
\usepackage{algorithm}
\usepackage{multirow}
\usepackage{svg}
\usepackage[normalem]{ulem}

\setcopyright{acmcopyright}
\copyrightyear{2024}
\acmYear{2024}

\begin{document}

\title{Trainable Fixed-Point Quantization for Deep Learning Acceleration on FPGAs}

\author{Dingyi Dai}
\affiliation{
    \institution{Cornell University}
    \country{}
}
\email{dnd29@cornell.edu}

\author{Yichi Zhang}
\affiliation{
    \institution{Cornell University}
    \country{}
}
\email{yz2499@cornell.edu}

\author{Jiahao Zhang}
\affiliation{
    \institution{Tsinghua University}
    \country{}
}
\email{jiahao-z19@mails.tsinghua.edu.cn}

\author{Zhanqiu Hu}
\affiliation{
    \institution{Cornell University}
    \country{}
}
\email{zh338@cornell.edu}

\author{Yaohui Cai}
\affiliation{
    \institution{Cornell University}
    \country{}
}
\email{yc2632@cornell.edu}

\author{Qi Sun}
\affiliation{
    \institution{Cornell University}
    \country{}
}
\email{qs228@cornell.edu}

\author{Zhiru Zhang}
\affiliation{
    \institution{Cornell University}
    \country{}
}
\email{zhiruz@cornell.edu}

\renewcommand{\shortauthors}{}

\begin{abstract}

Quantization is a crucial technique for deploying deep learning models on resource-constrained devices, such as embedded FPGAs. Prior efforts mostly focus on quantizing matrix multiplications, leaving other layers like BatchNorm or shortcuts in floating-point form, even though fixed-point arithmetic is more efficient on FPGAs. A common practice is to fine-tune a pre-trained model to fixed-point for FPGA deployment, but potentially degrading accuracy. 

This work presents QFX, a novel trainable fixed-point quantization approach that automatically learns the binary-point position during model training. Additionally, we introduce a multiplier-free quantization strategy within QFX to minimize DSP usage. QFX is implemented as a PyTorch-based library that efficiently emulates fixed-point arithmetic, supported by FPGA HLS, in a differentiable manner during backpropagation. With minimal effort, models trained with QFX can readily be deployed through HLS, producing the same numerical results as their software counterparts. Our evaluation shows that compared to post-training quantization, QFX can quantize models trained with element-wise layers quantized to fewer bits and achieve higher accuracy on both CIFAR-10 and ImageNet datasets. We further demonstrate the efficacy of multiplier-free quantization using a state-of-the-art binarized neural network accelerator designed for an embedded FPGA (AMD Xilinx Ultra96 v2). We plan to release QFX in open-source format.
\end{abstract}


  
\maketitle

\section{Introduction}
Quantization has been one of the primary techniques to improve the efficiency of deep neural network (DNN) inference. There is an active body of work focusing on quantizing the matrix multiplications in DNNs~\cite{banner2018scalable, migacz2017nvidia}. Layers such as batch normalization (BatchNorm), activation functions, and residual connections, however, typically remain floating-point during training~\cite{brevitas}. At the deployment time, on the other hand, FPGA accelerators usually convert these floating-point operations to fixed-point, since they are more efficient, through post-training quantization (PTQ)~\cite{yang2019synetgy, zhang2021fracbnn}. Eventually, there are two different models for training and FPGA-based inference.
However, PTQ has several drawbacks: (1) PTQ requires extensive finetuning on fixed-point precision. Practical FPGA DNN accelerators, for example, commonly have dozens of normalization layers, activation functions, or residual connections. Each of these layers may have a different fixed-point bitwidth in order to minimize the model accuracy loss compared to its floating-point counterpart and in the meantime optimize for fewer hardware resources. Therefore, finetuning the bitwidth is intractable due to the large design space. (2) PTQ obviously requires a larger word length, i.e., more hardware resources, than quantization-aware training (QAT) to maintain the same model accuracy.

This paper explores fixed-point quantization-aware training to address the aforementioned problems. 
Specifically, we introduce QFX, a differentiable \underline{q}uantized \underline{f}i\underline{x}ed-point emulation technique. QFX emulates the fixed-point casting function and basic arithmetic operations, e.g., addition, subtraction, and multiplication. All operations can properly propagate gradients and can be applied anywhere in a DNN during model training. At deployment time, the fixed-point operations in the model can be directly replaced by their synthesizable counterparts supported by HLS without any numerical issues. 
Therefore, the trained model is exactly the one that will be deployed on an FPGA.

\textbf{We leverage QFX and automatically learn the position of the binary point during model training.} Fixed-point data types are determined by two hyperparameters: word length and integer length. Given the word length, the integer length affects the range of a fixed-point value and is critical for DNN model accuracy. Instead of tuning the integer length manually, QFX can automatically learn it through model training for each layer where fixed-point quantization is applied. 

\textbf{We further propose a differentiable K-hot multiplier-free quantization scheme.} To minimize DSP usage on FPGAs, we develop a new quantization approach, where the fixed-point representation only includes a select few "1"s, allowing us to substitute multiplications with additions and shifts. We are able to apply this novel scheme to binarized neural networks (BNNs) to construct multiplier-free DNN accelerators on FPGAs.

\textbf{We demonstrate the efficacy of our proposed approach on both accuracy and resource usage.} We benchmark QFX against PTQ on several popular DNN models using both CIFAR-10 and ImageNet datasets. Our trainable quantization method consistently outperforms PTQ in accuracy, especially in low-bitwidth settings. We further show that the K-hot quantization scheme drastically reduces the DSP usage in FracBNN~\cite{zhang2021fracbnn}, a state-of-the-art BNN accelerator on an AMD Xilinx Ultra96 v2 FPGA. 



%

\section{Background}
In this section, we provide a concise introduction to the fundamental concepts of integer and fixed-point quantization, laying the groundwork for later discussions.

\subsection{Integer Quantization}
\label{sec:int-quant}

Earlier research in this field primarily focused on integer quantization, where DNNs' floating-point weights and activations are quantized to low-bitwidth integer values. This allows inference to be executed efficiently through hardware-accelerated integer matrix multiplications. 

Typically, quantization functions map a floating-point tensor $r$ to an integer tensor $q$ using the following formula: 
$$\texttt{quant}(r) = \texttt{round}(\texttt{clip}(\frac{r-Z}{S})).$$
Here,   $S$ denotes the scaling factor, and $Z$ serves as the zero point.
$S$ and $Z$ determines the range of values that can be represented by the quantized number. \texttt{clip} cuts the value outside the range and \texttt{round} operation maps floating-point value
to an integer.

After computations of each quantized layer, the output q will be dequantized to floating-point: 
$$\texttt{dequant}(q) = q \times S + Z.$$

\textbf{Fake quantization.} Fake quantization is often used to simulate QAT ~\cite{jacob2018quantization, brevitas}. In this approach, weights and activations are simulated as integers with floating-point representation, and all the multiplications and accumulations occur in FP32 so that the gradient computation is enabled. Nevertheless, HAWQ-V3~\cite{yao2021hawq} mentioned that this method results in significant error accumulation when deployed on real hardware, primarily due to the discrepant integer castings on the simulation and the actual hardware. 
 
\textbf{Scaling factor quantization.} Efforts~\cite{jacob2018quantization, shawahna2022fxp, yao2021hawq} have also been made to quantize the scaling factor to lower precision, recognizing that it is often directly cast to a fixed-point number during deployment on hardware which exacerbates the software-hardware gap. HAWQ-V3 proposed that we can construct the multiplier consisting of the scaling factors as a dyadic number during training. A dyadic number $b\cdot2^{-c}$ is essentially a fixed-point number, where $b$ is an integer, and $c$ represents the number of fractional bits. This approach not only replaces the multiplication between integer output and floating-point scaling factor with integer multiplication and bit shift but also helps avoid direct casting errors.

\subsection{Fixed-point Quantization}
While the computational overhead related to floating-point scaling factor calculations can be negligible compared to matrix multiplication in convolutional layers, element-wise layers themselves only involve one or two floating operations. Therefore, rather than representing a floating-point tensor as a low-bitwidth integer tensor and a floating-point scalar $S$,
an alternative approach is to quantize it to a fixed-point tensor, as fixed-point operations are also efficient on hardware ~\cite{shawahna2022fxp}. A floating-point tensor can be quantized to fixed-point with a given total bitwidth $w$ and an integer bitwidth $i$ through Algorithm~\ref{alg:fixed-point}.
\begin{algorithm}
\caption{Fixed-point quantization}\label{alg:fixed-point}
\begin{algorithmic}
\Function{Quant}{$x,w,i$}
    \State $fbit$ = $w - i$
    \State $\hat{x}$ = $x\cdot \text{pow}(2,fbit)$ \Comment{$x << fbit$}
    \State $\hat{x}$ = $\text{overflow}(\text{round}(\hat{x}), w)$
    \State $x$ = $\hat{x} \div \texttt{pow}(2,fbit)$ \Comment{$x >> fbit$}\\
    \Return $x$
\EndFunction
\end{algorithmic}
\end{algorithm}

Here, \texttt{round} operation rounds the number to nearest integer,  removing extra bits in the fractional part, and
\texttt{overflow} ensures the number falls into the representation range with $w$ bitwidth, removing extra bits in the integer part.


\section{Methodology}

In this section, we first introduce the differentiable fixed-point quantization emulation library (QFX), which enables fixed-point model training. We then describe how we leverage the library to automatically learn the fixed-point data type through model training. Finally, we demonstrate the construction of a multiplier-free CNN using these techniques on top of a BNN.

\subsection{QFX Implementation}
\label{sec:qfx-method}

\begin{figure}[t]
  \includegraphics[width=\columnwidth]{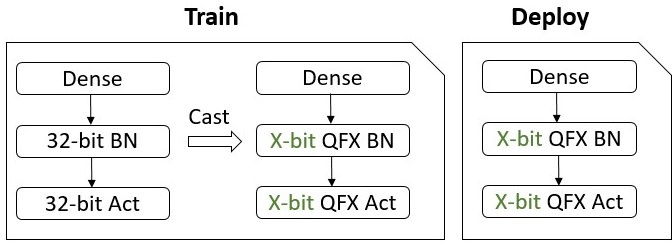}
  \caption{QFX design flow.\textnormal{ Element-wise operations are cast to fixed-point at training time without the ``quantization'' and ``dequantization'' overhead. After QAT, exactly the same model will be deployed on the target hardware, with the casting function replaced by hardened circuits.}}
  \label{fig:qfx}
\end{figure}

The fixed-point emulation in our QFX library is encapsulated in regular PyTorch layers inheriting \verb|nn.module|. The implementation is all through native PyTorch tensor operations, which makes QFX layers easily be plugged into existing PyTorch models. Most importantly, QFX layers are backpropagation compatible. Models with QFX quantization can support normal gradient descent training.

An example flow of using the QFX library is shown in Figure~\ref{fig:qfx}. In contrast to the traditional practice where there are different model copies for training and deployment, users can apply the fixed-point casting, addition, or multiplication operations anywhere in a given PyTorch model at training time. Exactly the same model will be deployed, except that the data type casting will be replaced directly by the hardened circuits. Below we introduce how fixed-point casting and basic arithmetic are implemented and how they are assembled into a quantized BatchNorm layer as an example.

\textbf{Fixed-point casting.}
Given a floating-point input $x$, the following equation converts it to a fixed-point representation:
\begin{equation*}
    \operatorname{cast}(x, wbit, fbit) = \text{overflow}(\text{round}(x << fbit), wbit) >> fbit.
\end{equation*}
For the quantization function, rounding towards the nearest integer is the most common case. On FPGAs, however, various circuits break the tie differently. QFX supports seven quantization modes, the implementation of which is shown in Table~\ref{table:quant-mode}. It produces the same numerical outputs as the reference AMD Xilinx \verb|ap_fixed| library~\cite{xilinx2022vitis}. Similarly, all available overflow implementations are shown in Table~\ref{table:overflow-mode}. One may notice that the casting function leveraging PyTorch native operators is inherently differentiable except for \verb|floor|, \verb|ceil|, \verb|round|, and \verb|trunc| functions. Their gradients are almost zero everywhere. To enable training with QFX, we redefine their gradients as:
$$\frac{\partial \operatorname{floor}\left(x\right)}{\partial x} = \frac{\partial \operatorname{ceil}\left(x\right)}{\partial x} = \frac{\partial \operatorname{round}\left(x\right)}{\partial x} = \frac{\partial \operatorname{trunc}\left(x\right)}{\partial x} := 1,$$
which is known as the straight-through estimator (STE)~\cite{bengio2013estimating}. STE is commonly used to approximate gradients for non-differentiable operations~\cite{mishra2017wrpn, 2020CVPRW-LSQ+, 2019CVPR-OptInterval}.
Users can also define customized gradient functions and integrate them under the QFX training scheme.

Supporting a wide range of commonly used quantization and overflow modes guarantees that the model inference during software training and FPGA deployment are aligned. However, DNN models are typically trained on GPUs or CPUs where rounding to the nearest even number, i.e., RND\_CONV in Table~\ref{table:quant-mode}, is supported by default. This subtle difference, especially in a DNN accelerator~\cite{zhang2021fracbnn}, will lead to completely different hardware outputs. In order to synchronize the outputs produced by both software quantization emulation and FPGA deployment, the general QFX approach is necessary.


\textbf{Basic arithmetic.}  
QFX supports a subset of fixed-point element-wise operations that are the most common in DNNs, including addition, subtraction, multiplication, and division. To enable fixed-point arithmetic during model training, we apply the casting function to both the input and output of these operations. For example, the floating-point addition operation:
\begin{equation*}
    \operatorname{add}\left( x_1, x_2 \right) = x_1 + x_2,
\end{equation*}
transforms into the following for fixed-point computation within QFX:
\begin{equation*}
    \operatorname{qfx.add}\left( x_1, x_2 \right) = \operatorname{cast}\left( \operatorname{cast}\left(x_1\right) + \operatorname{cast}\left(x_2\right) \right).
\end{equation*}
Each \verb|cast| function has its own word length and fractional bitwidth. Importantly, the basic operations remain differentiable with QFX casting functions. When deploying the model, the casting functions can be seamlessly replaced by circuits, by programming the data type of input and output variables using high-level synthesis (HLS) or hardware programming languages. The other three operations --- subtraction, multiplication, and division, follow a similar approach.

\begin{table}[t]
  \centering
  \caption{Supported round modes $\texttt{round}(x)$ in QFX .}
  \label{table:quant-mode}
  \begin{tabular}{|l|l|}
    \hline
    \textbf{Quantization Mode} & \textbf{Implementation}\\
    \hline
    RND           & $\operatorname{torch.floor}(x + 0.5)$ \\
    \hline
    RND\_ZERO     & $\begin{cases} \operatorname{torch.floor}\left ( x+0.5 \right ) & \text{ if } x < 0 \\ \operatorname{torch.ceil}\left ( x-0.5 \right ) & \text{ if } x \ge 0 \end{cases}$ \\
    \hline
    RND\_MIN\_INF & $\operatorname{torch.ceil}\left( x-0.5 \right)$ \\
    \hline
    RND\_INF      & $\begin{cases} \operatorname{torch.ceil}\left ( x-0.5 \right ) & \text{ if } x < 0 \\ \operatorname{torch.floor}\left ( x+0.5 \right ) & \text{ if } x \ge 0 \end{cases}$ \\
    \hline
    RND\_CONV \tablefootnote{Also referred to as round to the nearest even.}     & $\operatorname{torch.round}\left( x \right)$ \\
    \hline
    TRN           & $\operatorname{torch.floor}\left( x \right)$ \\
    \hline
    TRN\_ZERO     & $\operatorname{torch.trunc}\left( x \right)$ \\
    \hline
  \end{tabular}
\end{table}

\begin{table}[t]
  \centering
  \caption{Supported overflow modes $\texttt{overflow}(x, wbit)$ in QFX. }
  \label{table:overflow-mode}
  \begin{tabular}{|l|l|}
    \hline
    \textbf{Overflow Mode} & \textbf{Implementation}\tablefootnote{Signed and unsigned are treated equivalently. For signed values, $\text{min\_val}=-2^{wbit-1}$ and $\text{max\_val}=2^{wbit-1}-1$. For unsigned fixed-point values, $\text{min\_val}=0$ and $\text{max\_val}=2^{wbit}-1$.}\\
    \hline
    SAT & $\operatorname{torch.clamp}\left ( x, \text{min\_val}, \text{max\_val} \right )$ \\
    \hline
    SAT\_ZERO & $\begin{cases} x & \text{ if } \text{min\_val} \le x \le \text{max\_val} \\ 0 & \text{ otherwise} \end{cases}$ \\
    \hline
    SAT\_SYM\_SIGNED & $\operatorname{torch.clamp}\left ( x, -\text{max\_val}, \text{max\_val} \right )$ \\
    \hline
    SAT\_SYM\_UNSIGNED & $\operatorname{torch.clamp}\left ( x, 0, \text{max\_val} \right )$ \\
    \hline
    WRAP\_SIGNED \tablefootnote{$\rho = \text{max\_val} - \text{min\_val} + 1$} & $\begin{cases} x \mod \rho & \text{ if } x \mod \rho \le \text{max\_val} \\ x \mod \rho - \rho & \text{ if } x \mod \rho > \text{max\_val} \end{cases}$ \\
    \hline
    WRAP\_UNSIGNED & $x \mod \rho$ \\
    \hline
  \end{tabular}
\end{table}

\textbf{Quantized element-wise operations.}
With the supported basic arithmetic, common element-wise operations, e.g., BatchNorms and residual connections, can be emulated in fixed-point during training.
As an example, the floating-point BatchNorm layer was computed by:
\begin{equation*}
    \operatorname{BN} \left( x \right) = \frac{x - \mathbb{E}\left[x\right]}{\sqrt{\operatorname{Var}\left(x\right) + \epsilon}} \cdot \gamma + \beta,
\end{equation*}
where $\beta$ and $\gamma$ are learnable parameters, and $\epsilon$ is a small value for numerical stability. With fixed-point quantization, it becomes:
\begin{equation*}
    \operatorname{qfx.BN} \left( x \right) = \operatorname{qfx.add} \left( \operatorname{qfx.mul} \left( \alpha, x \right), \eta \right),
\end{equation*}
where $\alpha = \frac{\gamma}{\sqrt{\operatorname{Var}\left(x\right) + \epsilon}}$ and $\eta = \beta - \frac{\gamma \cdot \mathbb{E}\left[x\right]}{\sqrt{\operatorname{Var}\left(x\right) + \epsilon}}$. Note that the casting operation is embedded within \verb|qfx.add| and \verb|qfx.mul|. Both input and output can be specified with different word lengths and fractional bitwidths. The QFX library includes pre-implemented fixed-point BatchNorm layers and residual connections, which are commonly used in  DNNs. Users can easily customize their models  to fixed-point by substituting the four basic operations with QFX.

\subsection{Learning Binary Point Position}
In Section~\ref{sec:qfx-method}, we have introduced the QFX library.
In this section, we delve into how QFX can be harnessed to automate the learning of fixed-point data types, thereby eliminating the need for extensive precision tuning in practical applications.

A fixed-point representation can be divided into two parts:  the integer and the fraction, separated by the binary point.
The number of bits allocated to the integer and fractional parts determines the range and precision of the fixed-point representation.
For example, a decimal number $(13.3125)_d$ can be accurately represented by a fixed binary number $(1101.0101)_b$, with $4$ integer bits and $4$ fractional bits:
$$\underset{\text{Int=4}}{\underline{1101}}.\underset{\text{Frac=4}}{\underline{0101}}.$$

Previous work~\cite{loroch2017tensorquant, abadi2016tensorflow, shawahna2022fxp} in fixed-point quantization primarily focused on fine-tuning PTQ with the binary point position fixed, demanding additional human effort to strike the right balance between range and precision.

In our work, QFX offers support for fixed-point QAT with different configurations for each layer and even for different weights within a single layer. Consequently, automating the binary point position searching is as straightforward as assigning a new parameter to Algorithm  \ref{alg:fixed-point} with gradient backpropagation enabled. In this case, we are able to improve the accuracy with all quantization effects included during training. 

\begin{algorithm}
\begin{flushleft} 
\caption{Automate binary-point search}\label{alg:binary-point}
\hspace*{0.1in}{\bf Init parameters:} \\
\hspace*{0.2in} \textbf{Set} integer\_bits $I$, grad=True\\
\hspace*{0.1in}{\bf forward:}\\
\hspace*{0.2in} $\hat{I}$ = \text{round}(\text{clamp}($I, 0, w$))\\
\hspace*{0.2in} $x$ = Quant($x, w, \hat{I}$)
\end{flushleft}
\end{algorithm}

As shown in Algorithm \ref{alg:binary-point}, the trainable floating point $I$ is rounded to integer to be fed into the fixed-point casting function of $x$.
Notably, gradient backpropagation in \texttt{torch.clamp} is enabled via STE, and QFX also enables gradient backpropagation for rounding. This ensures that the function is differentiable and binary point position remains trainable. 

\subsection{K-hot Fixed-point Quantization}
Dedicated DSP units on FPGAs are often considered as a limited resource in contrast to the ample availability of LUTs. Even for highly compressed DNN models, a nontrivial number of DSPs are typically still required for operations using fixed-point arithmetic. By profiling a state-of-the-art BNN accelerator, FracBNN, we find that more than 60\% of the DSPs on its Ultra96v2 FPGA deployment are still in use, even if all the matrix multiplications in the model are binarized. This presents a suboptimal scenario for edge devices. To mitigate this and further conserve DSP resources, this section introduces a multiplier-free fixed-point quantization scheme, which incurs zero DSP usage. 

Instead of solely quantizing the weights in element-wise layers (e.g., BatchNorms) to fixed-point format, we further quantize them to a K-hot encoding --- enforcing it to have only $K$ "1"s in its binary representation:
\begin{align}
x_K &= \sum_{i=1}^K 2^{K_i - fbit}  \\
&= \sum_{i=1}^K \text{\text{bshift}}(1, K_i - fbit)
\end{align}
where $K_i$ is the position of the $i$-th most significant "1", $fbit$ is the fractional bits, and \texttt{bshift} shifts $1$ left by $K_i - fbit$ bits in binary format. One multiplication between a fixed-point number $x_f$ and an integer number $x_i$ typically requires DSP units. By imposing the $K$-hot constraint with a small $K$, the multiplication is replaced by at most $K$ additions and bit shifts.
\begin{align}
x_K \cdot x_i &= \sum_{i=1}^K \text{\text{bshift}}(1, K_i - fbit) \cdot x_i \\
&= \sum_{i=1}^K \text{\text{bshift}}(x_i, K_i - fbit )
\end{align}
This $K$-hot fixed-point leads to an entirely \textbf{DSP-free} design.

To convert a floating-point input to a k-hot fixed-point number, we first use the Quant casting function in QFX to quantize it to fixed-point with corresponding integer bits $I$. Then we find the $K$ most significant bits to approximate the number in binary representation. $K$ can be flexibly configured based on the accuracy-efficiency trade-off. A smaller $K$ is more efficient but less accurate. Detailed algorithm description can be found in Algorithm \ref{alg:two-hot}.

\begin{algorithm}
\caption{K-hot Quantization}\label{alg:two-hot}
\begin{algorithmic}
\State \textbf{Init parameters:}
    \State \hspace*{0.15in} \textbf{Set} integer\_bits $I$, grad=True
\Function{K\_hot}{$x, w, k$}
    \State $\hat I$ = \text{round}(\text{clamp}($I, 0, w$))
    \State $qfx\_x$ = Quant($x, w, \hat I$)
    \State $\hat{x}$ = 0
    \For{$i$ in range($k$)}
        \State $k_i$ = \text{floor}(\text{log2}($qfx\_x - \hat{x}$)
        \State $\hat{x}$ += $\text{bshift}(1, k_i)$
    \EndFor \\
    \Return $\hat{x}$
\EndFunction
\end{algorithmic}
\end{algorithm}



\section{Evaluation}
In this section, we first evaluate the efficacy of QFX on accuracy using four representative convolutional neural networks (CNNs), including ResNet-18~\cite{he2016deep}, ResNet-50~\cite{he2016deep}, MobileNetV2~\cite{sandler2018mobilenetv2}, and EfficientNet-B0~\cite{tan2019efficientnet}. Then we present QFX results on three popular BNNs, including FracBNN-CIFAR-10~\cite{zhang2021fracbnn},  Bi-Real Net \cite{liu2018bi}, and FracBNN-ImageNet. FracBNN-CIFAR-10 is trained on the CIFAR-10 dataset~\cite{krizhevsky2009learning}, while the others are trained on the ImageNet dataset~\cite{deng2009imagenet}.  We compare them with PTQ and show QFX achieves considerable gains in accuracy. 
Additionally, we present the results of applying 2-hot quantization to the BNNs, which make them multiplier-free. We design an HLS accelerator for FracBNN-ImageNet to evaluate the hardware performance when employing these techniques.

\subsection{Evaluation of Model Accuracy}
We quantize element-wise operations to fixed-point, e.g., BatchNorm layers, activations functions, and residual connections. Both input activations and layer weights are quantized to target bitwidth, with the binary point position of each quantized number being automatically learned during training. Subsequently, we employ 2-hot quantization in BNNs for all the weight multipliers in BatchNorm layers and activation functions. 

\paragraph{Quantization configurations.} In the experiments, we use RND/ TRN\_ZERO rounding mode and SAT overflow mode since they are commonly used in quantization. 
For the baseline PTQ, we sweep binary point positions to find the configuration that minimizes the mean square error (MSE) of each layer output, which represents the gap between the floating-point layers and post-training fixed-point quantized layers.

\paragraph{Training details.} The training strategy for QFX can vary depending on the model architecture, model size, and training cost. For the CNNs, we only fine-tune for 5 epochs, since they are more resilient to low bitwidth element-wise operation with the convolutional layers remaining in floating-point. The training is executed on NVIDIA RTX 2080Ti GPUs, employing a learning rate of 1e-5 and a batch size of 64. 

For FracBNN-CIFAR-10, we follow the two-step training strategy defined in FracBNN and train the model from scratch. QFX fixed-point quantization is applied at the second training step. For the other two BNN models, we only fine-tune QFX quantization for 30 epochs from the pretrained checkpoint due to the long training time. The hyperparameters are the same as defined in their original papers except that the learning rate of FracBNN-ImageNet is adjusted linearly with batch size of 128 due to limited GPU memory. We further fine-tune models with 2-hot quantization for 10 more epochs with a 10$\times$ smaller learning rate. Further fine-tuning may lead to a higher accuracy but at a higher training cost. The training for BNNs are conducted on NVIDIA RTX A6000 GPUs. All the experiments use the Adam optimizer~\cite{kingma2014adam} without weight decay. 

\paragraph{Accuracy Results.}
Table~\ref{tab:nonBNN} reports the accuracy results on CNN models. QFX outperforms PTQ significantly at 8 bits, and has similar accuracy at higher bits. This clearly shows the advantages of emulating fixed-point arithmetic during training and providing the support of differentiable quantization functions.

Table~\ref{tab:QAT accuracy} shows the results of BNNs. With a larger bitwidth, e.g., 16 bits, PTQ can easily obtain good accuracy, especially on a small dataset. However, when quantized to 8 bits, PTQ is unable to produce reasonable accuracy, while QFX can still preserve the accuracy. For FracBNN-CIFAR-10, quantizing to 8 bits using QFX only causes a marginal 0.59\% loss in accuracy compared to the BNN with floating-point element-wise layers. For Bi-Real Net, the loss is only 1.23\% on ImageNet. Unlike other BNNs where the first convolutional layers are in high-precision, FracBNN has all convolutional layers binarized and appears to be more sensitive to the precision of element-wise layers. 
Nevertheless, it still has 69.27\% accuracy at 8 bits with QFX. 

Table~\ref{tab:multiplier-less accuracy} shows the results of applying 2-hot quantization on the multipliers within element-wise layers. After only 10 epochs of fine-tuning, we achieve an accuracy loss of less than 0.4\% on both datasets with FracBNN. Notably, the slight improvement for Bi-Real Net underscores the potential for QFX to yield even better results with extended fine-tuning, indicating the possibility of lossless performance with multiplier-free quantization. 


\begin{table}
  \caption{Top-1 accuracy (\%) of CNNs with element-wise layers quantized to fixed-point using PTQ \& QFX at 10/8 bits (W = total bitwidth).}
  \label{tab:nonBNN}
  \begin{tabular}{lr|crr}
    \toprule
    Network&FP32&Method & W=10 & W=8\\
    \midrule
    \multirow{2}{*}{ResNet-18}& \multirow{2}{*}{69.9} & PTQ & 66.10  & 42.78  \\
    & & QFX& \textbf{69.86} & \textbf{67.86} \\
    \hline
    \multirow{2}{*}{ResNet-50}& \multirow{2}{*}{76.1}& PTQ & 47.03& 0.12\\
    & & QFX& \textbf{75.67}& \textbf{67.80}\\
    \hline
 \multirow{2}{*}{MobileNetV2}& \multirow{2}{*}{72.2}& PTQ & 8.11&0.14\\
 & & QFX&\textbf{ 70.19}&\textbf{57.45}\\
 \hline
 \multirow{2}{*}{EfficientNet-B0}& \multirow{2}{*}{76.1}& PTQ & 74.87&69.90\\
 & & QFX& \textbf{76.92}&\textbf{76.26}\\
    \bottomrule
  \end{tabular}
\end{table}

\begin{table}
  \caption{Top-1 accuracy (\%) of BNNs with element-wise layers quantized to fixed-point using PTQ \& QFX at 16/12/10/8 bits (W = total bitwidth).}
  \label{tab:QAT accuracy}
  \begin{tabular}{lr|crrrr}
    \toprule
    Network&FP32&Method& W=16 & W=12 & W=10 & W=8\\
    \midrule
    FracBNN-& \multirow{2}{*}{88.7} & PTQ & \textbf{88.70}& 87.40 & 80.70 & 11.90 \\
    CIFAR-10& & QFX& 88.34 & \textbf{88.63} & \textbf{88.65} & \textbf{88.11} \\
    \hline
    \multirow{2}{*}{Bi-Real Net}& \multirow{2}{*}{56.4} & PTQ & 54.90 & 45.50 & 7.90 & 0.40 \\
    & & QFX& \textbf{55.75} & \textbf{55.60} & \textbf{55.68} & \textbf{55.17}\\
    \hline
    FracBNN-& \multirow{2}{*}{71.8} & PTQ & 69.70 & 23.30 & 0.70 &  0.10 \\
    ImageNet& & QFX& - & - & - &  \textbf{69.27}\\
    \bottomrule
  \end{tabular}
\end{table}

\begin{table}
  \caption{Top-1 accuracy (\%) of fixed-point models  using QFX w/ and w/o 2-hot encoding at 8 bits.}
  \label{tab:multiplier-less accuracy}
  \begin{tabular}{l|rcr}
    \toprule
    Network  &QFX & QFX w/ 2-hot & Diff \\
    \midrule
    FracBNN-CIFAR-10 &88.11 & 87.89 & -0.22  \\
    Bi-Real Net &55.17 & 55.31 & 0.14 \\
    FracBNN-ImageNet &69.27 & 68.95 & -0.32  \\
    \bottomrule
  \end{tabular}
\end{table}

\subsection{FPGA Implementation}



\begin{figure}
    \centering
    \includegraphics[width=0.85\linewidth]{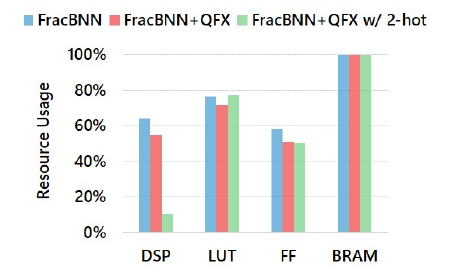}
    \caption{Resource usage on an AMD Xilinx Ultra96 v2 FPGA.}
    \label{fig:resource}
\end{figure}

\begin{figure}
    \centering
    \includegraphics[width=0.9\linewidth]{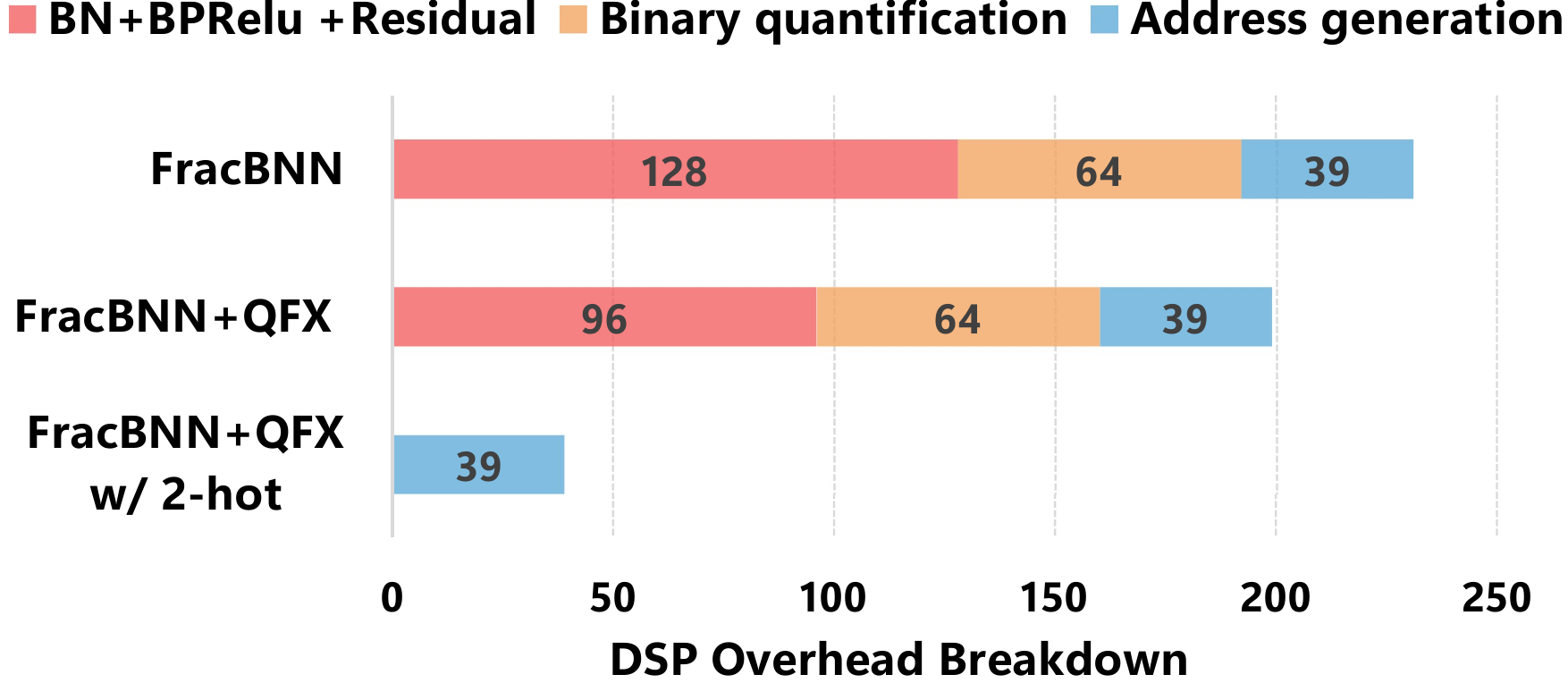}
    \caption{DSP breakdown w/ different quantization methods.}
    \label{fig:DSP}
\end{figure}

\begin{table}
  \caption{Resource usage of element-wise layers when adopting 8-bit QFX w/ and w/o 2-hot quantization or forcing the fabric implementation using \textit{BIND\_OP} pragma in HLS.}
  \label{tab:Bn_Relu_Residual overheads}
  \begin{tabular}{l|rrr}
    \toprule
    Resource & QFX & QFX w/ 2-hot & \textit{BIND\_OP imp=fabric}  \\
    \midrule
    DSP &128 &0 (100\%$\downarrow$) &0 (100\%$\downarrow$) \\
    FF &7222 &9034 (25.1\%$\uparrow$) &15497 (114.6\%$\uparrow$) \\
    LUT &2944 &3676 (24.9\%$\uparrow$) &4421 (50.2\%$\uparrow$) \\
  \bottomrule
\end{tabular}
\end{table}

To evaluate the performance and resource improvements by QFX, we implement our FracBNN-ImageNet accelerator with 8-bit QFX quantization on an AMD Xilinx Ultra96 v2 FPGA board, which has 360 DSPs, 71k LUTs, 141K FFs and 949 KB BRAMs. We preserve the existing binary and fractional convolution layers in the original FracBNN accelerator. Our efforts focus on applying QFX to the element-wise layers (including BatchNorm, BPRelu, and residual), as well as binary quantization layers, thereby achieving a DSP-free design in these regions without any degradation in throughput.

With the $ap\_fixed$ library provided by HLS, it is straightforward to cast all the weights and activations of different element-wise layers to 8-bit fixed-point numbers upon obtaining the learned data types from the training stage. In contrast, the fixed-point data types used in the original FPGA implementation were determined manually.  
Additionally, our K-hot quantization technique facilitates  DSP-free implementation in BatchNorm and BPRelu layers. In our design, both layers can be represented as:
\begin{equation*}
    y=\hat{a}\cdot x+b,
\end{equation*}
where $\hat{a}$ represents a two-hot quantized fixed-point parameter with only two 1s. Thus we can easily transform the multiplication into simple bit shift and addition. 
Besides replacing DSPs with 2-hot MAC units in element-wise layers, similar optimizations apply to binary quantization layers, where scaling factors could be quantized using K-hot encoding. 

Figure \ref{fig:resource} compares the hardware performance of our 8-bit QFX accelerators against the original FracBNN accelerator. By limiting the bitwidth of fixed-point numbers in both element-wise and binary quantization layers, 8-bit QFX decreases DSP, LUT and FF overheads by 8.9\%, 4.7\%, and 7.0\%,  respectively. Moreover, the multiplier-free design can substantially reduce DSP usage from 55.3\% to 10.8\%, while only increasing 5.8\% and 1.1\% LUT overhead compared to 8-bit QFX without 2-hot encoding and the original FracBNN design.

Figure \ref{fig:DSP} further provides a breakdown of the DSP usage. We observe that 8-bit QFX reduces total DSP usage from 231 to 199, while the 2-hot encoding completely eliminates DSP usage in element-wise layers and binary quantization layers. This leaves only 39 DSPs, which are used for memory address calculations. 

Notably, compared to simply using the \textit{BIND\_OP imp=fabric} pragma in HLS to avoid instantiating DSPs, our design is much more efficient. As shown in Table \ref{tab:Bn_Relu_Residual overheads}, using the HLS \textit{fabric} pragma to remove DSPs leads to 114.6\% and 50.2\% LUT and FF increases for element-wise layers with 8-bit QFX quantization. In contrast, the multiplier-free technique with 2-hot encoding requires much fewer LUTs and FFs.

\section{Related Work}
\textbf{Integer quantization on matrix multiplication.} Early work focused on 8-bit uniform quantization \cite{banner2018scalable, holt1993finite, migacz2017nvidia} as it was widely accessible on commercial hardware such as CPUs and GPUs. As domain-specific ML hardware emerged \cite{jouppi2021ten}, researchers started to explore lower precision quantization. Recently, binarized neural networks (BNNs) have demonstrated competitive performance \cite{qin2023bibench, yuan2023comprehensive} in terms of accuracy-efficiency trade-off compared to float ones on both vision and language tasks \cite{zhang2023binarized, zhang2022pokebnn}. Our work is motivated by the profiling on FracBNN.


\textbf{Quantization-aware training.} To improve quantized model accuracy further, QAT is necessary \cite{abdolrashidi2021pareto, 2020CVPRW-LSQ+, 2019CVPR-OptInterval}. The key is enabling gradients to propagate effectively "through" the quantization function, allowing the network to account for the quantization loss, even though its gradients are theoretically zero at all points. To tackle the challenge, various gradient approximation techniques have been proposed, including the Straight-Through Estimator (STE) \cite{bengio2013estimating}, differentiable soft tanh \cite{gong2019differentiable}, and even using Fourier transforms \cite{xu2021learning}. In our QFX implementation, we achieve differentiability by leveraging the STE, thereby supporting QAT.

\textbf{Fixed-point emulation} An existing framework QNNPACK~\cite{dukhan2018qnnpack} optimizes fixed-point quantization for mobile devices but only with 8 bits. TensorQuant~\cite{loroch2017tensorquant} and Tensorflow~\cite{abadi2016tensorflow} emulate arbitrary
bitwidth fixed point but lack of support on multiple rounding and overflow modes. QPyTorch~\cite{2019-EMC2-NIPS-QPyTorch} has three rounding modes while its casting is not differentiable. Brevitas~\cite{brevitas} supports most rounding modes but lacks fixed-point quantization aware training for element-wise layers.. Typically, existing quantization training tools on software cover  a subset of quantization modes or lack differentiability to facilitate QAT.

\section{Conclusion}
This work introduces QFX, a new approach to  trainable fixed-point quantization. QFX is implemented as a PyTorch library, which can emulate different quantization modes provided by FPGA HLS tools and enables backpropagation for quantization parameters. QFX can automate the process of binary-point determination, enabling a systematic search of the optimal precision level for each layer.  
We further introduce the K-hot quantization to transform fixed-point multiplications into a series of bit shifts and additions. This design, combined with binary multiply-accumulate operations in BNNS, culminates in a ``DSP-free'' approach, which can be beneficial for deploying compressed deep learning models on embedded FPGAs.


\bibliographystyle{ACM-Reference-Format}
\bibliography{reference}


\end{document}